\documentclass[lettersize,journal]{IEEEtran}

\usepackage[switch]{lineno} % switch allows line numbers on the left column in two-column mode
\usepackage{etoolbox}       % for patching commands

\makeatletter
\patchcmd{\@outputdblcol}
  {\global\@linesperpage}
  {\let\linelabel\@gobble  \global\@linesperpage}
  {}{}
\makeatother

\usepackage{amsmath,amsfonts}
\usepackage{algorithmic}
\usepackage{algorithm}
\usepackage{array}

\usepackage{textcomp}
\usepackage{stfloats}
\usepackage{url}
\usepackage{verbatim}
\usepackage{graphicx}
\usepackage{multirow}
\usepackage{booktabs}
\usepackage{tabularx}
\usepackage{cite}

\usepackage{caption}  
\usepackage{subcaption} 
\usepackage{hyperref}

\begin{document}

\title{Neural ATTF: A Scalable Solution to Lifelong Multi-Agent Path Planning}

\author{Kushal Shah$^1$, Jihyun Park$^2$, Seung-Kyum Choi$^1$\\
$^1$Georgia Institute of Technology, Atlanta, GA, USA \\
$^2$Hyundai-MOBIS, Seoul, Korea}

% The paper headers
% \markboth{Journal of \LaTeX\ Class Files,~Vol.~14, No.~8, August~2021}%
% {Shell \MakeLowercase{\textit{et al.}}: A Sample Article Using IEEEtran.cls for IEEE Journals}

% \IEEEpubid{0000--0000/00\$00.00~\copyright~2021 IEEE}
% Remember, if you use this you must call \IEEEpubidadjcol in the second
% column for its text to clear the IEEEpubid mark.

\maketitle
% \linenumbers
\begin{abstract}
Multi-Agent Pickup and Delivery (MAPD) is a fundamental problem in robotics, particularly in applications such as warehouse automation and logistics. Existing solutions often face challenges in scalability, adaptability, and efficiency, limiting their applicability in dynamic environments with real-time planning requirements. This paper presents Neural ATTF (Adaptive Task Token Framework), a new algorithm that combines a Priority Guided Task Matching (PGTM) Module with Neural STA* (Space-Time A*), a data-driven path planning method. Neural STA* enhances path planning by enabling rapid exploration of the search space through guided learned heuristics and ensures collision avoidance under dynamic constraints. PGTM prioritizes delayed agents and dynamically assigns tasks by prioritizing agents nearest to these tasks, optimizing both continuity and system throughput.%The heuristic-driven task assignment module is a centralized module that dynamically allocates tasks to agents based on a heuristic, optimizing both task completion times and overall system throughput. 
Experimental evaluations against state-of-the-art MAPD algorithms, including TPTS, CENTRAL, RMCA, LNS-PBS, and LNS-wPBS, demonstrate the superior scalability, solution quality, and computational efficiency of Neural ATTF. These results highlight the framework's potential for addressing the critical demands of complex, real-world multi-agent systems operating in high-demand, unpredictable settings. 
\end{abstract}


\section{Introduction}
\IEEEPARstart{M}{any} real-world applications of multi-agent systems require agents to operate in shared environments while dynamically executing tasks. These tasks often involve ensuring collision-free trajectories as agents move between locations to fulfill their objectives. Examples include aircraft towing vehicles \cite{morris2016planning}, warehouse robots \cite{wurman2008coordinating}, and office service robots \cite{veloso2015cobots}. This class of problems can be formalized as the Multi-Agent Pickup and Delivery (MAPD) problem, which consists of two interrelated subproblems: (1) assigning agents to pickup-and-delivery tasks, and (2) planning collision-free paths for each agent to ensure successful task execution. Both task assignment and multi-agent pathfinding are computationally intractable challenges on their own \cite{korsah2013comprehensive,stern2019multi}. The online nature of MAPD—where tasks are generated continuously and are unknown at the outset—further compounds the complexity, making efficient and scalable solutions highly challenging to achieve.

Several approaches have been proposed in the recent literature to address the Multi-Agent Path Finding (MAPF) and MAPD problem. Algorithms such as Conflict-Based Search with Task Assignment (CBS-TA) \cite{honig2018conflict} provide optimal solutions for MAPF but are limited in scalability, as they perform effectively only in small environments and struggle with a large number of agents or expansive workspaces. A prominent algorithm, LNS-wPBS \cite{xu2022multi}, combines Large Neighborhood Search (LNS) for task assignment with windowed Priority-Based Search (PBS) for path planning. While this method demonstrates improved performance in certain scenarios, the windowed nature of the path planner can result in deadlocks, limiting its robustness in dynamic settings. Another notable algorithm is Regret-based Marginal Cost-based Task Assignment (RMCA) \cite{chen2021integrated}, a coupled approach where task assignment and path planning are performed simultaneously. By utilizing actual cost instead of heuristics for task assignment, RMCA improves efficiency and solution quality. However, this comes at the cost of increased computational effort, leading to scalability challenges in larger environments or with a greater number of agents.

% In this work, we introduce the Neural Adaptive Task Token Framework (Neural ATTF), a decoupled algorithm designed to address the MAPD problem efficiently and scalably. Neural ATTF solves the task assignment problem using a heuristic-driven task assignment module and addresses the path planning problem through a space-time extension of the Neural A* algorithm. For path planning, Neural ATTF integrates the principles of Space-Time A* and Neural A*. A guidance map provides heuristic information to direct the path planning process, significantly reducing computational overhead while ensuring collision-free trajectories. Through extensive simulations, we demonstrate that Neural ATTF is more scalable and more efficient than current state-of-the-art algorithms such as TPTS \cite{ma2017lifelong}, CENTRAL \cite{ma2017lifelong}, HBH-MLA* \cite{grenouilleau2019multi}, RMCA \cite{chen2021integrated}, LNS-PBS, and LNS-wPBS \cite{xu2022multi}. Additionally, we extend Neural ATTF to handle unexpected delays and uncertainties in agent operations, enhancing its robustness in dynamic and unpredictable environments. These capabilities make Neural ATTF a robust and scalable solution to the MAPD problem.

In this work, we introduce the Neural Adaptive Task Token Framework (Neural ATTF), a decoupled algorithm designed to address the Multi-Agent Pickup and Delivery (MAPD) problem with high efficiency, scalability, and robustness. Neural ATTF solves the task assignment problem using a heuristic-driven assignment module and tackles path planning through a space-time extension of Neural A*, which combines the temporal modeling of Space-Time A* with the data-driven efficiency of Neural A*. The framework introduces several novel contributions. First, it proposes a globally-informed Priority Guided Task Matching (PGTM) Module that prioritizes delayed agents and dynamically assigns tasks based on lightweight distance heuristics, thereby prioritizing agents nearest to these tasks and optimizing both continuity and system throughput. Second, it presents the Neural STA* (Space-Time A*) path planner, a hybrid algorithm that efficiently computes collision-free trajectories by leveraging learned heuristics in the space-time domain. Third, Neural ATTF incorporates a novel idling and deadlock recovery mechanism that avoids indefinite waiting at endpoints by rerouting idle agents to safe non-task locations or nearby free cells, preventing gridlock and improving flow. Lastly, it extends its framework to explicitly handle execution delays and uncertainties, allowing agents to replan and adapt in real time, thereby increasing system resilience in dynamic and unpredictable environments. Through extensive simulations, we demonstrate that Neural ATTF outperforms state-of-the-art algorithms such as TPTS \cite{ma2017lifelong}, CENTRAL \cite{ma2017lifelong}, HBH-MLA* \cite{grenouilleau2019multi}, RMCA \cite{chen2021integrated}, LNS-PBS \cite{xu2022multi}, and LNS-wPBS \cite{xu2022multi} in both scalability and efficiency, making it a strong candidate for real-time deployment in complex multi-agent systems.

\section{Preliminaries and Background}
An instance of the MAPD problem consists of two subproblems: task assignment and pathfinding. In this section, we provide an overview of existing research in these areas and introduce Neural A* \cite{yonetani2021path}, which will be employed as the low-level planner to compute collision-free paths for agents.

\subsection{Task Assignment}
The task assignment aspect of the MAPD problem shares significant overlap with multi-robot task allocation and vehicle routing problems. Nguyen et al. \cite{nguyen2019generalized} tackled a generalized target assignment and pathfinding problem using a three-phase approach tailored to a simplified warehouse setting. \IEEEpubidadjcol 
However, their method introduced unnecessary waiting between phases and demonstrated limited scalability, handling only up to 20 tasks or robots. Multi-robot task allocation has been widely explored \cite{korsah2013comprehensive,bai2020event}, with notable approaches including the Hungarian algorithm \cite{kuhn1955hungarian}, a combinatorial optimization method that efficiently finds maximum-weight matchings in bipartite graphs. A prominent local search technique is Large Neighborhood Search (LNS) \cite{shaw1997new}, which addresses the vehicle routing problem by iteratively refining an initial schedule. In each iteration, a subset of agents is removed based on a removal heuristic and reinserted, potentially at different positions, using a greedy heuristic. Additionally, pickup-and-delivery problems, such as the dial-a-ride problem \cite{cordeau2007dial}, have been extensively studied, with heuristic assignment algorithms and regret-based methods \cite{tillman1972upperbound,potvin1993parallel,zheng2008agent} proposed to optimize task assignment under various constraints.

\subsection{Multi-agent Path Finding}
The Multi-Agent Path Finding (MAPF) problem, an essential part of MAPD, has been widely studied with approaches categorized into optimal, bounded-suboptimal, prioritized, and rule-based solvers. Optimal solvers like Conflict-Based Search (CBS) \cite{sharon2015conflict} and Branch-and-Cut-and-Price (BCP) \cite{bettinelli2011branch} provide guarantees of optimality but struggle with scalability. Bounded-suboptimal solvers such as Enhanced CBS (ECBS) \cite{cohen2016improved} balance scalability with near-optimality. Prioritized solvers, including Cooperative A* (CA) \cite{silver2005cooperative} and Priority-Based Search (PBS) \cite{ma2019searching}, plan paths based on agent priorities, though they can be incomplete or suboptimal for complex scenarios. Rule-based solvers like Parallel Push and Swap \cite{sajid2012multi} offer polynomial-time solutions, though their quality is lower. Multi-Label A* (MLA*) \cite{grenouilleau2019multi} and its extension handle multi-goal pathfinding, such as in pickup and delivery tasks. While these methods have improved efficiency, scalability remains a challenge, especially when integrated with task assignment in large MAPD problems. Additionally, windowed MAPF solvers, such as windowed PBS, employ the Rolling-Horizon Collision Resolution (RHCR) technique \cite{li2021lifelong} to plan paths for large numbers of agents in the MAPD problem. However, these methods assume task assignment is handled by a separate system and are prone to deadlocks due to their short-sightedness.

\subsection{Multi-agent Pickup and Delivery}
The MAPD problem has been tackled using both centralized and decentralized approaches. Token Passing \cite{ma2017lifelong} is a decentralized algorithm that solves MAPD online using a method similar to Cooperative A*. In contrast, the CENTRAL algorithm \cite{ma2017lifelong} uses a centralized approach, applying the Hungarian method for task assignment and Conflict-Based Search (CBS) for path planning, suitable for dynamic, online settings. TA-Hybrid \cite{liu2019task} is an offline solution that models task assignment as a Traveling Salesman Problem (TSP) and uses CBS for path planning, though both methods struggle with scalability. These approaches are decoupled, assigning tasks first and then planning paths with an MAPF solver. RMCA \cite{chen2021integrated}, a coupled approach, enhances efficiency by using Large Neighborhood Search (LNS) for task assignment and prioritized planning with A* for pathfinding. Additionally, heuristic-based methods like H-value-based Heuristic (HBH) have been explored, followed by prioritized planning and Multi-Label A* for path planning \cite{grenouilleau2019multi}. Recent work also includes LNS-PBS and LNS-wPBS \cite{xu2022multi}. LNS-PBS combines LNS for task assignment with Priority-Based Search (PBS) and ``reserving dummy paths" for path finding, emphasizing completeness and effectiveness. In contrast, LNS-wPBS improves efficiency and stability by integrating the windowed PBS with RHCR \cite{li2021lifelong} to handle larger MAPD instances.

\subsection{Some Existing MAPD Algorithms} 
\subsubsection{TP and TPTS}
Token Passing (TP) \cite{ma2017lifelong} is a decentralized algorithm for multi-agent task assignment and path planning. In TP, a token circulates among agents, allowing only one agent at a time to make decisions. When an agent receives the token, it considers unassigned tasks in order of increasing heuristic cost to their pickup locations. If a task is unassigned, the agent assigns itself, plans a collision-free path, updates the token with this path, and returns success. If no suitable task is found, the agent either plans a trivial resting path or moves to a designated endpoint to avoid deadlocks. Completed tasks are removed from the task set as agents execute them.

Token Passing with Task Swaps (TPTS) \cite{ma2017lifelong} extends TP by allowing dynamic task reassignment. Instead of considering only unassigned tasks, TPTS includes all unexecuted tasks in the task set. This enables an agent holding the token to reassign itself to a task already assigned to another agent, as long as the originally assigned agent is still en route to the pickup location. If the current agent can reach the pickup location faster, it unassigns the other agent, updates the token with its own path, and passes the token to the unassigned agent for a new task search. If the reassignment leads to overall improvement, it is retained; otherwise, changes are reverted, and the next task is considered. This mechanism enhances adaptability and promotes more efficient use of agents by reducing idle time and allowing opportunistic task optimization.

\subsubsection{CENTRAL}
The CENTRAL algorithm \cite{ma2017lifelong} is a centralized approach to solving the Multi-Agent Pickup and Delivery (MAPD) problem. It operates in discrete timesteps, assigning endpoints to all agents and planning collision-free paths for them simultaneously. In each timestep, CENTRAL first assigns tasks to agents that are at the pickup locations of unexecuted tasks, provided their delivery locations are unassigned. These agents become occupied and start executing their assigned tasks. For free agents, CENTRAL assigns either the pickup location of an unexecuted task or a parking location to ensure all endpoints remain unique. To construct these endpoints, CENTRAL considers only valid pickup and delivery locations from tasks that avoid conflicts with other agents and, if necessary, adds parking locations based on proximity. The Hungarian Method is used to assign endpoints to agents, prioritizing pickup locations over parking spots while minimizing the overall cost of assignments. Once endpoints are assigned, CENTRAL uses Conflict-Based Search (CBS) to plan optimal collision-free paths for all agents. Paths are planned in two stages to improve efficiency: first for occupied agents and then for free agents, treating the paths of other agents as spatiotemporal obstacles. This staged approach reduces computational complexity and ensures the solution respects constraints. Agents then move along their assigned paths for one timestep, and the process repeats. By combining centralized coordination with optimized path planning, CENTRAL achieves high effectiveness and serves as a benchmark to evaluate the performance of algorithms like Token Passing (TP) and Token Passing with Task Swaps (TPTS).

\subsubsection{RMCA}
The Regret-based Marginal Cost-based Task Assignment (RMCA) \cite{chen2021integrated} algorithm is an efficient approach to solving the MAPD problem by integrating task assignment and path planning simultaneously. Instead of assigning tasks first and planning paths later, RMCA selects tasks using a regret heuristic, prioritizing those where the cost difference between the best and second-best robot assignments is highest. Each robot maintains an ordered sequence of pickup and delivery actions, and tasks are assigned while updating paths dynamically using prioritized space-time A* search to ensure collision-free movement. RMCA continuously refines solutions through anytime improvement strategies, such as randomly reassigning tasks, removing poorly performing assignments, or redistributing multiple tasks across robots. Designed for real-time, lifelong MAPD scenarios, RMCA efficiently adapts to new tasks while minimizing total travel delay, allowing robots to complete tasks sooner and remain available for new assignments. By leveraging real-time path costs instead of lower-bound estimates, RMCA improves task assignments, reduces computational complexity compared to centralized methods like CBS-TA, and enhances system throughput, making it highly effective for applications like warehouse automation and logistics.

\subsubsection{HBH-MLA*}
The HBH-MLA* algorithm \cite{grenouilleau2019multi} is a centralized approach for solving the Multi-Agent Pickup and Delivery (MAPD) problem, integrating an h-value-based heuristic (HBH) with a Multi-Label A* (MLA*) path planner. Unlike traditional A*, which computes paths sequentially (first to the pickup location and then to the delivery location), MLA* simultaneously searches for both, reducing unnecessary constraints and allowing agents to move efficiently. HBH assigns tasks iteratively based on the h-value, prioritizing agents with the shortest estimated travel distance to a task. At each timestep, idle agents are assigned to available tasks using MLA*, ensuring collision-free paths. If no tasks are available, idle agents are moved to nearby free locations to avoid blocking others. This combined approach improves both computation time and solution quality, significantly reducing makespan and service time compared to existing methods.

\subsubsection{LNS-PBS}
LNS-PBS \cite{xu2022multi} is a decoupled algorithm designed to solve the Multi-Agent Pickup and Delivery (MAPD) problem by combining Large Neighborhood Search (LNS) for task assignment with Priority-Based Search (PBS) for path planning. LNS iteratively improves task sequences assigned to agents by first constructing an initial solution using a Hungarian-based insertion heuristic and then refining it through Shaw removal and regret-based re-insertion. The algorithm assigns each agent a sequence of tasks while ensuring that all goal locations remain distinct from assigned dummy endpoints—locations where agents can remain indefinitely without interfering with others. PBS is then used to compute collision-free paths for agents by dynamically resolving priority conflicts through a depth-first search in a priority tree. To ensure completeness for well-formed MAPD instances, LNS-PBS enforces constraints that prevent deadlocks and guarantees that all agents reach their assigned goal locations in finite time. This makes LNS-PBS highly effective for structured environments, offering a balance between computational efficiency and solution quality while maintaining provable completeness guarantees.

\subsubsection{LNS-wPBS}
LNS-wPBS \cite{xu2022multi} is a variant of LNS-PBS that optimizes computational efficiency and scalability by integrating windowed PBS (wPBS) instead of standard PBS. Instead of computing complete paths to all assigned goal locations, LNS-wPBS plans paths for a fixed time horizon ($w$ timesteps) and replans once agents have moved for $w$ steps, significantly reducing computational complexity. Unlike LNS-PBS, LNS-wPBS does not defer tasks based on endpoint conflicts, simplifying task assignment and enabling continuous execution without requiring strict endpoint constraints. Additionally, wPBS does not consider past paths when replanning, further improving computational efficiency. Although LNS-wPBS does not guarantee completeness due to its limited planning horizon, it demonstrates empirical robustness in large-scale MAPD instances with thousands of agents and tasks, outperforming existing scalable algorithms like HBH+MLA* in terms of service time while maintaining significantly lower computational overhead.

\subsection{Neural A*}
Neural A* \cite{yonetani2021path} is a data-driven search method for path planning that combines a convolutional encoder with a differentiable version of the A* search algorithm. It reformulates the traditional A* search to be differentiable, making it compatible with end-to-end trainable neural network planners. The method encodes the problem instance, including the environmental map and start/goal points, into a guidance map using a convolutional encoder. This map is then used by the differentiable A* module to perform the search. By training to match the search results with ground-truth paths from traditional and optimal planners like A* and Dijkstra, Neural A* generates paths that are both accurate and efficient. The differentiable nature of Neural A* allows backpropagation of losses through each search step, enabling the encoder to learn effective visual cues for path planning, thereby improving search optimality and efficiency.

\section{Problem Formulation}

Consider an instance of the MAPD problem that consists of a set of $n$ agents \( A = \{a_1, a_2, \dots, a_n\} \), which operate within an undirected, connected graph \( G = (V, E) \). Here, \( V \) represents the set of vertices corresponding to locations, and \( E \) represents the set of edges that indicate possible movement routes between locations. Each agent \( a_i \) is initially located at a vertex \( v_i(0) \in V \), and at each discrete time step \( t \), the agent occupies a vertex \( v_i(t) \in V \). The agent’s movement is restricted in that it can either remain at its current vertex or move to an adjacent vertex along an edge, such that \( (v_i(t), v_i(t+1)) \in E \) or \( v_i(t+1) = v_i(t) \). Importantly, the movement of agents must avoid collisions, both spatial and temporal.

Each task \( \tau_k \) is defined by a pickup location \( p_k \in V \), a delivery location \( d_k \in V \), and a release time \( r_k \), after which it is added to the unassigned task set \( T \). Agents can only be assigned tasks after their release time. A free agent, not engaged in any task, can be assigned a task \(\tau_k\), requiring it to travel from its current location to \( p_k \), and then proceed to \( d_k \). Once an agent is assigned the task \(\tau_k\), it is removed from \( T \), and upon reaching \( d_k \), the agent completes the task and becomes free. In practical warehouse scenarios, an agent must complete the entire task it is assigned before starting another. Here, the pickup and delivery locations $p_k$ and $d_k$ are referred to as task endpoints, as they serve as critical waypoints that define the start and completion of a task.

The objective is to minimize the average service time, which is the duration between task arrival and task completion. Formally, the service time for a task $\tau_k$ assigned to agent $a_i$ is defined as:
\[
    t_{service} = \frac{1}{m}\sum_{k=0}^m(\min \{t | v_i(t) = d_k, x_{ik} = 1\} - r_k)
\]
where 
\begin{equation*}
  x_{ik} =
    \begin{cases}
      1 & \text{if agent $a_i$ assigned to task $\tau_k$}\\
      0 & \text{otherwise,}
    \end{cases}       
\end{equation*}
$m$ is the total number of tasks, $r_k$ is the time step when task $\tau_k$ is released in the system, and $t_{service}$ represents the time taken for the agent to complete the task after assignment. Hence, the final optimization problem is stated as:

\begin{figure}
    \centering
    \includegraphics[width=0.7\linewidth]{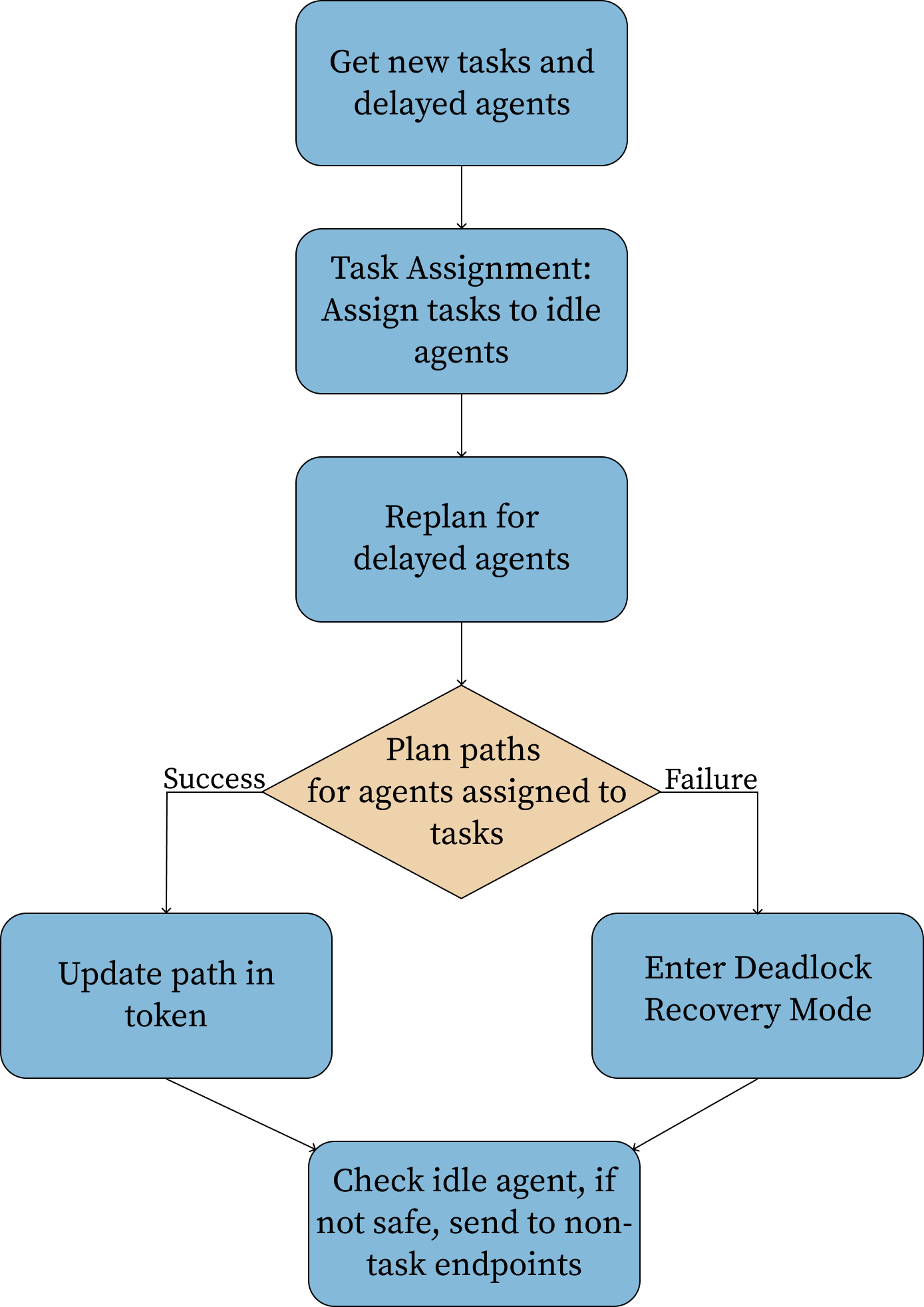}
    \caption{Overall flow of Neural ATTF in each timestep}
    \label{fig:flow}
\end{figure}

\begin{align}
   \text{Minimize} \;\;\;& t_{service} & \notag \\
   \text{subject to} \;\;\; & v_i(t) \neq v_j(t)\;\; \forall i \neq j, \; \forall t, \; \forall a_i, a_j \in A\\ 
   & (v_i(t), v_i(t+1)) \neq (v_j(t+1), v_j(t)) \;\; \forall i \neq j, \notag \\
   & \forall t, \; \forall a_i, a_j \in A\\
   & (v_i(t), v_i(t+1)) \in E \; \; \forall a_i \in A,\; \forall t\\
   & v_i(t) \in V \;\; \forall a_i \in A,\;\forall\;t\\
   & x_{ik}(t) = 1 \implies v_i(t_1) = p_k , v_i(t_2) = d_k \notag \\
   & t \in [t_1, t_2], \; \forall a_i \in A,\; \forall \tau_k \in T \\
   & \text{if} \;\; x_{ik}(t) = 1, v_i(t) = p_k \implies \exists\;t' \geq t \notag \\
   & \text{s.t.}\;\; v_i(t') = d_k \;\; \forall a_i \in A,\;\forall \tau_k \in T \\
   & x_{ik}(t) = 1 \;\; \text{only if} \;\; t > r_k \;\; \forall a_i \in A, \forall \tau_k \in T \\
   & \sum_{k\in T}x_{ik}(t) \leq 1 \;\; \forall a_i \in A,\; \forall t\\
   & \sum_{a_i \in A}x_{ik}(t) \leq 1 \;\; \forall \tau_k \in T, \;\forall t
\end{align}

% \begin{align}
    
% \end{align}

Here (1) defines the vertex collision constraint and (2) defines the edge collision constraint. The constraints (3) and (4) restrict the robots to node set \(V\) and the movement of robots to the edges \(E\) of the graph \(G\). Constraint (5) denotes that the task will be executed by an agent if it is assigned to it. Constraint (6) requires the same agent to drop off the task if it picks up the task. Constraint (7) ensures that the task is not assigned before its release time. Constraint (8) make sure that an agent takes up no more than one task at a time. (9) ensures that a task is assigned to no more than 1 agent at a time. 

\subsection*{Delays}

Delays are common in real-world MAPF and MAPD due to factors like sensor errors, physical constraints (e.g., turning radius, velocity, acceleration) \cite{ma2019lifelong}, and anomalies like partial failures during execution. While algorithms can convert time-discrete MAPD plans into executable ones \cite{ma2019lifelong}, mismatches between models and actual robots still cause delays. Delays are modeled as agents remaining in their current positions instead of moving forward on their planned paths, with these delays being unpredictable during planning but finite to ensure well-formedness \cite{lodigiani2023robust}.

\begin{algorithm}[t]
\caption{Neural ATTF}
\label{alg:algorithm1}
\begin{algorithmic}[1]
    \STATE Initialize token, path for each agent $a_i = [v_i(0)]$
        \WHILE{True} 
        \STATE Update completed tasks
        \STATE ${D} \gets$ Delayed Agent Set
        \STATE Add all new tasks to the task set $T$
        \STATE $T' \gets \{\tau_k \in T | \tau_k\;\; \text{not assigned to any agent}\}$
        \STATE ${IdleAgents}$ $\gets$ All agents not assigned to a task
        \STATE $AssignedPairs \gets Assign(D, IdleAgents, T')$
        \STATE $Env \gets$ Current State with idle agent obstacles
        \STATE $CostMaps \gets$ Encoder($Env$, $AssignedPairs$)
        \WHILE{$len(IdleAgents) > 0$}
            \STATE $a_i, \tau_k \gets AssignedPairs.pop(0)$
            \IF {$\tau_k$ is not None}
                \STATE $costmap \gets CostMaps.pop(0)$ 
                \STATE Remove $a_i$ from $IdleAgents$
                \STATE Mark $\tau_k$ as assigned and remove from $T'$
                \STATE $token[a_i] = Plan(Loc(a_i), \tau_k(s), costmap)$
                \STATE $token[a_i] \mathrel{{+}{=}} Plan(\tau_k(s), \tau_k(g), costmap)$
            \ELSE
                \IF {$a_i$ safe at current location}
                    \STATE \textbf{continue}
                \ELSE
                    \STATE $l_{ntp} \gets \text{arg min}_{e_i \in N}\;h(loc(a_i), e_i)$
                    \STATE $token[a_i] = Plan(a_i, l_{ntp})$
                    \STATE Remove $a_i$ from $IdleAgents$
                \ENDIF
            \ENDIF
        \ENDWHILE
        \STATE Agents move along their paths for one time step

        \ENDWHILE
\end{algorithmic}
\end{algorithm}

\section{Neural ATTF}

The proposed algorithm is a robust and efficient algorithm, with a simple Priority Guided Task Matching (PGTM) Module, and a guided space-time A* driven path planning module. A simplified flow of the algorithm is shown in figure \ref{fig:flow}. It operates in two key stages: task assignment and path planning. It is designed for robust, real-time coordination in dynamic multi-agent environments. It first uses a centralized, globally-informed heuristic to assign tasks to idle and delayed agents, prioritizing agents that are delayed to preserve their task assignment, and then agents that are closest to task pickup locations. Once tasks are assigned, the algorithm sequentially plans paths for each agent in a decentralized manner following the order received from the task assignment module. For each agent, we construct a spatial grid of the environment to generate guidance maps using a neural encoder. These serve as inputs to the path planning module (Space-Time A*) that computes efficient, collision-free paths. Idle agents without tasks are either allowed to stay put if safe or rerouted to non-task endpoints to avoid obstructing others. This tight integration of adaptive task assignment, learning-based planning, and intelligent idling ensures scalability and robustness under uncertainty.

The overall algorithm is shown in algorithm \ref{alg:algorithm1}. We store all the agent paths and task assignments in the token, similar to TP. At each timestep, we first update the completed tasks [line 3]. Then we check if any agent is delayed and add it to a set $D$ [line 4]. Then we collect new tasks added to the system at this timestep, get the unassigned task set $T'$ and get all idle agents in the system [line 5-7]. We then assign tasks to each of the idle and delayed agents using the task assignment module explained in section \ref{subsec:ta}.  

After assigning tasks to the idle agents, we construct a grid map representing the current state of the environment, marking obstacles and idle agents as obstacles (for that particular timestep) [line 9]. Additionally, for each assigned agent, we generate two one-hot encoded binary maps: the first marks the agent's current position and the task's start location, and the second marks the task's start and goal locations. These map pairs capture the spatial relationships relevant to the agent’s task. For $k$ assigned agents, this results in $2k$ individual maps. All maps are batched together into a tensor of shape $[2k, H, W]$, where $H$ and $W$ are the height and width of the map. This batch is then processed in parallel by a shared convolutional encoder (described in Section \ref{subsec:enc}), which produces one guidance map for each input. As a result, each agent receives two guidance maps: one guiding the path from its current position to the task's start location, and the other guiding the path from the task's start to its goal location.

Next, we iterate through every idle agent and access its assigned task from the queue. If there is an assigned task for the agent, we get the corresponding guidance map for that task, and plan the path from current location to task start and then to task goal location using space-time neural A* algorithm [line 17 -18]. This algorithm is explained in detail in \ref{subsec:sta*}. Else, the algorithm verifies whether the agent is safe at its current location and does not obstruct any active agents. If the location is safe, the agent is allowed to idle there [line 20-21]. Otherwise, a path is planned to send the agent to the nearest non-task endpoint, ensuring it remains out of the way while waiting for new tasks [line 23-25]. Once we iterate through all the agents, all the agents move along their paths for one timestep.

A key improvement in this algorithm is the relaxation of the idling constraint for agents at the end of their paths. Traditional algorithms like TP and TPTS sequentially plan paths, assuming agents rest indefinitely at path ends, which over-constrains the problem \cite{ma2017lifelong}. MLA* addressed this partially by planning a complete path from the agent's current location to the task goal via the task start \cite{grenouilleau2019multi} but still assumed indefinite idling at the goal. Our approach eliminates this constraint, by assuming that the agents idle at the path end for only one timestep. If not assigned a new task immediately, the agent's safety is checked for idling at its position. This one-timestep buffer ensures the agent is not treated as an obstacle at the path end, except when it is at its non-task endpoint, enabling other agents to plan around it effectively and reducing delays and conflicts.

\begin{figure*}[t]
    \centering
    \includegraphics[width=\linewidth]{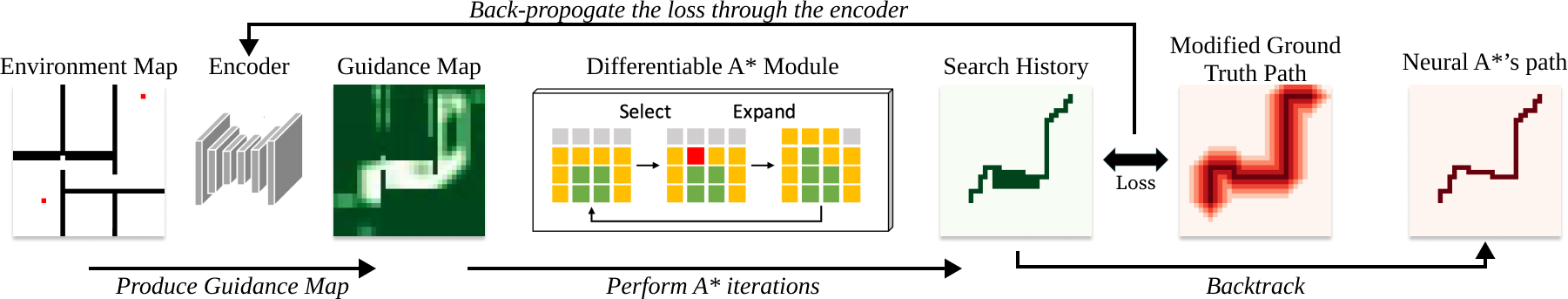}
    \caption{Training Pipeline of Neural A*. The environment is encoded to get a guidance map, which is used by differentiable A* module to generate the shortest path and search history. A loss between the search history and the ground-truth path is back-propagated to train the encoder.}
    \label{fig:neurala}
\end{figure*}

\subsection{Priority Guided Task Matching Module}
\label{subsec:ta}
We employ a globally-informed, heuristic-driven task assignment module that prioritizes delayed agents and then assigns tasks by prioritizing agents based on proximity. We initialize a first-in-first-out queue by first adding delayed agents along with their current tasks to preserve the original task assignment. For each idle agent, we evaluate all available tasks by calculating a heuristic value for each agent-task pair. The heuristic is based on the Manhattan distance, with a small constant added to the Euclidean distance for tie-breaking between the agent's current location and the task start. These pairs are then sorted based on the heuristic value in ascending order. We ensure that agents and tasks are not duplicated in the queue by adding only unique agent-task pairs, with the lowest heuristic values prioritized for assignment. This order ensures that delayed agents are re-planned first, followed by agents that can reach the start location of their goal the fastest, optimizing the overall planning efficiency. If all tasks have been assigned, we assign $None$ to the remaining idle agents.

\subsection{Map Encoder}
\label{subsec:enc}
The encoder is a fully convolutional U-Net architecture \cite{ronneberger2015u} with a VGG-16 backbone, where the final layer is activated by a sigmoid function to constrain the output values between 0 and 1. This encoder produces the guidance maps, which are subsequently fed into the differentiable A* \cite{yonetani2021path} module to generate paths. The heuristic used in the A* module combines the Manhattan distance and a small constant ($0.001$) times the Euclidean distance for tie-breaking. In the original work, an L1 loss was backpropagated between the ground-truth path and the map of explored nodes during the search. We introduce a minor modification to account for the dynamic and constrained nature of our environment. Instead of relying solely on the ground-truth path, we weight cells inversely proportional to the Chebyshev distance from the path. This adjustment trains the network to assign relatively lower guidance values to minor deviations from the ideal path, which is critical for handling cases where some neighbors are invalid due to environmental constraints. This network was trained on the motion planning dataset \cite{bhardwaj2017learning}, which contains 8 different environment groups with distinct obstacle configurations, each comprising 800 training, 100 validation, and 100 test samples. These map groups include a wide variety of obstacle shapes and layouts, allowing the network to learn strong visual cues for diverse environments and improving its ability to generalize across unseen scenarios. To further enhance generalization, each data point was randomly resized to a resolution between $32 \times 32$ and $200 \times 200$. Since the encoder requires input dimensions to be multiples of 16, the maps were padded (with obstacles) to the nearest multiple of 16 in both height and width. The overall training pipeline is illustrated in Figure~\ref{fig:neurala}.

\begin{figure}
    \centering
    \includegraphics[width=\linewidth]{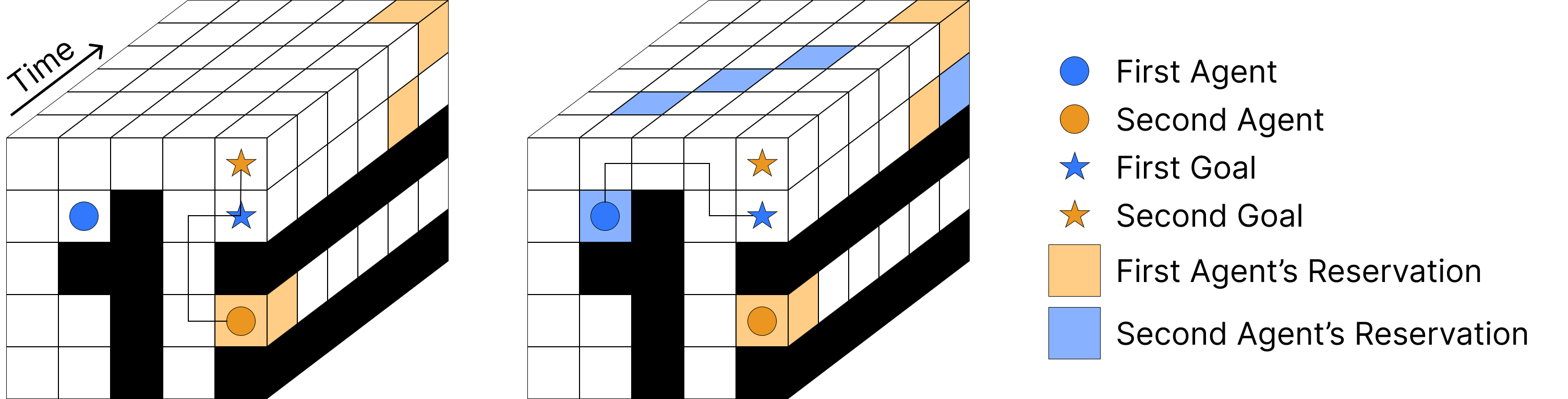}
    \caption{Illustration of Neural ATTF’s space-time path planning using Neural Space-Time A*. It depicts the 3D space-time planning grid, where each agent’s planned trajectory is shown along the time axis. Neural STA* uses dynamic occupancy maps to avoid collisions across time, enabling coordinated, collision-free navigation in multi-agent scenarios.}
    \label{fig:sta*_simple}
\end{figure}

\subsection{Neural Space-time A*}
\label{subsec:sta*}

This algorithm performs time-aware path planning by searching over both space and time, allowing agents to avoid collisions with other moving agents by treating their occupied positions as dynamic obstacles. It follows the general structure of A* search but extends it into the space-time domain, while using a learned cost map to guide expansion—making it a hybrid between classical and learned planning. The intuition is that, at each timestep, the algorithm explores the most promising future position (in space and time) by balancing the true cost to reach a node with an estimate of the remaining cost, while also avoiding regions blocked by other agents’ planned paths. A small problem is demonstrated in figure \ref{fig:sta*_simple}.

The psuedocode for the planning algorithm is shown in algorithm \ref{alg:algorithm2}. This algorithm is a variation of the Space-Time A* search method incorporating characterstics of Neural A*. This algorithm begins by initializing the cost-to-come ($g$) and estimated total cost ($f$) for the start location, where $f=g+h$, and $h$ is the heuristic function estimating the cost to the goal. The heuristic used is the Manhattan distance, as the environment is a 4-connected grid, with an additional tie-breaking term given by the Euclidean distance multiplied by a small constant ($0.001$). A priority queue, $Open$, manages nodes yet to be explored, and $Closed$ tracks visited locations. While the number of iterations do not exceed a user specified number of maximum iterations and $Open$ is not empty, the algorithm iteratively selects and removes the node with the lowest f-value from $Open$. If this node matches the goal, the path is reconstructed and returned. 

\begin{algorithm}[t]
\caption{Function $Plan$}
\label{alg:algorithm2}
\begin{algorithmic}[1]
\STATE \textbf{Input:} $Start, Goal, costmap$
\STATE $f[Start] = h(Start, Goal)$, $g[Start] = 0$
\STATE $Open = PriorityQueue([\;])$, $Closed = [\;]$
\STATE Push $(f[Start], h(Start, Goal), Start)$ to $Open$
\WHILE{$Open \neq \phi$ and Iterations $<$ MaxIters}
    \STATE Current $(x_i, y_i, t_i) \gets$ Pop top element of $Open$
    \IF{Current = $Goal$}
        \STATE Reconstruct path using parents
        \STATE \textbf{return} Path
    \ENDIF
    \STATE Add $Current$ to $Closed$
    \STATE $Neighbors \gets$ Valid neighbors of $Loc(Current)$ at time $Time(Current)$
    \FOR{$n$ in $Neighbors$}
        \STATE $tentative\:g = g[Current] + 1$
        \IF{$n \in Closed$ \AND $tentative\:g \geq g[n]$} 
            \STATE \textbf{continue}
        \ENDIF
        \IF{$tentative\:g < g[n]$}
            \STATE $n.parent = current$
            \STATE $g[n] = tentative\:g$
            \STATE $f[n] = g[n] + h(n, Goal) + costmap(n)$
            \STATE Push $(f[n], h(n, Goal), n)$ to $Open$
        \ENDIF
    \ENDFOR
\ENDWHILE
\STATE \textbf{return} Goal unreachable, deadlock recovery mode.
\end{algorithmic}
\end{algorithm}

Else, we add $Current$ to $Closed$ and process the valid neighbors of the current node. These neighbors are determined based on the current environment state and exclude locations occupied by other agents at that specific time. The neighbors can include the four adjacent grid cells or the current cell itself, representing a wait action. For each neighbor, a tentative cost-to-come is calculated. If the neighbor, has been visited earlier with a lower or equal cost, it is skipped. Otherwise, it is added to the search with updated $g$- and $f$-values, which incorporate the heuristic and costmap values to guide the expansion. This process repeats until the goal is reached or the maximum iterations are exceeded. In cases where an idle agent cannot find a valid path to a non-task endpoint—either due to temporary congestion or spatial limitations—it enters a deadlock recovery mode. 

\subsubsection*{Deadlock Recovery Mode}
In this mode, the system identifies a small local neighborhood around the agent (typically within a fixed radius $r$), and randomly selects a free cell that is not currently occupied or reserved by another agent's planned path. A short, collision-free path is then planned to this free cell using a constrained version of the Neural Space-Time A* planner, which uses a reduced planning horizon to minimize computational overhead. This redirection serves two key purposes: (1) it helps vacate congested areas or task-critical locations (e.g., pickup and delivery endpoints), and (2) it prevents agents from becoming permanent obstacles. As other agents execute their plans and space frees up in subsequent timesteps, the previously stuck agent re-enters the standard assignment and planning loop. This mechanism ensures forward progress and robustness in densely populated or dynamically blocked environments, effectively mitigating long-term gridlock and enabling continuous task execution.\\ \\
We now prove that Neural ATTF is complete for well-formed MAPD instances, following the style of the proof in \cite{ma2017lifelong}.\\ \\
\textbf{Property 1.}
The Neural Space-Time A* planner returns a collision-free path for well-formed MAPD instances.\\ \\
\textit{Proof.}
% In Neural ATTF, paths are planned sequentially, and each agent plans with full knowledge of other agents’ space-time trajectories stored in the global token. When planning a path for an agent $a_i$ from its current location to the pickup $p_k$ and delivery $d_k$ locations of task $\tau_k$, the Neural Space-Time A* planner ensures that the path avoids conflicts with all existing plans in the token. Since there always exists a path between any pair of endpoints that avoids traversing other endpoints \cite{ma2017lifelong} in well-formed instances, such a collision-free path always exists. Hence, the path returned by Neural STA* is valid.\\ \\ 
In well-formed MAPD instances \cite{ma2017lifelong}, there exists a path between any pair of endpoints that avoids traversing other endpoints. In Neural ATTF, paths are planned sequentially with full knowledge of all previously planned agent trajectories stored in the global token. When planning a path for an agent $a_i$ from its current location to the pickup $p_k$ and delivery $d_k$ of task $\tau_k$, the Neural Space-Time A* algorithm searches for a collision-free path in the space-time domain. Unlike traditional approaches that require agents to end at distinct endpoints, Neural ATTF allows multiple agents to use the same task endpoints across different times. Safety is maintained by ensuring that agents hold these endpoints for only one timestep before either being reassigned or redirected to a safe non-task endpoint or free cell (See property 2). This temporal staggering ensures that endpoints are eventually vacated and remain accessible to other agents in future planning steps. Therefore, each agent can safely execute its task without permanent conflicts, guaranteeing a valid path exists.\\ \\
\textbf{Property 2.}
Idle agents are eventually assigned safe, collision-free paths that prevent blocking of task endpoints.\\ \\
\textit{Proof.}
When no task is available for an agent, Neural ATTF attempts to either safely idle the agent at its current location (if it does not interfere with the paths of other agents or conflict with endpoints of unassigned tasks) or redirect it to a non-task endpoint. In well-formed instances, there are at least $m$ non-task endpoints for $m$ agents, ensuring that there is always at least one such endpoint not targeted by any other agent. If a valid path to such an endpoint exists, it is planned using Neural STA*. If not, the algorithm invokes a deadlock recovery mechanism that assigns a short, collision-free path to a nearby unoccupied cell, ensuring the agent vacates task-relevant regions. Therefore, agents do not indefinitely block task pickup or delivery locations.\\ \\

\begin{figure*}[t]
    \centering
    \begin{subfigure}{\linewidth}
        \centering
        \includegraphics[width=0.75\linewidth]{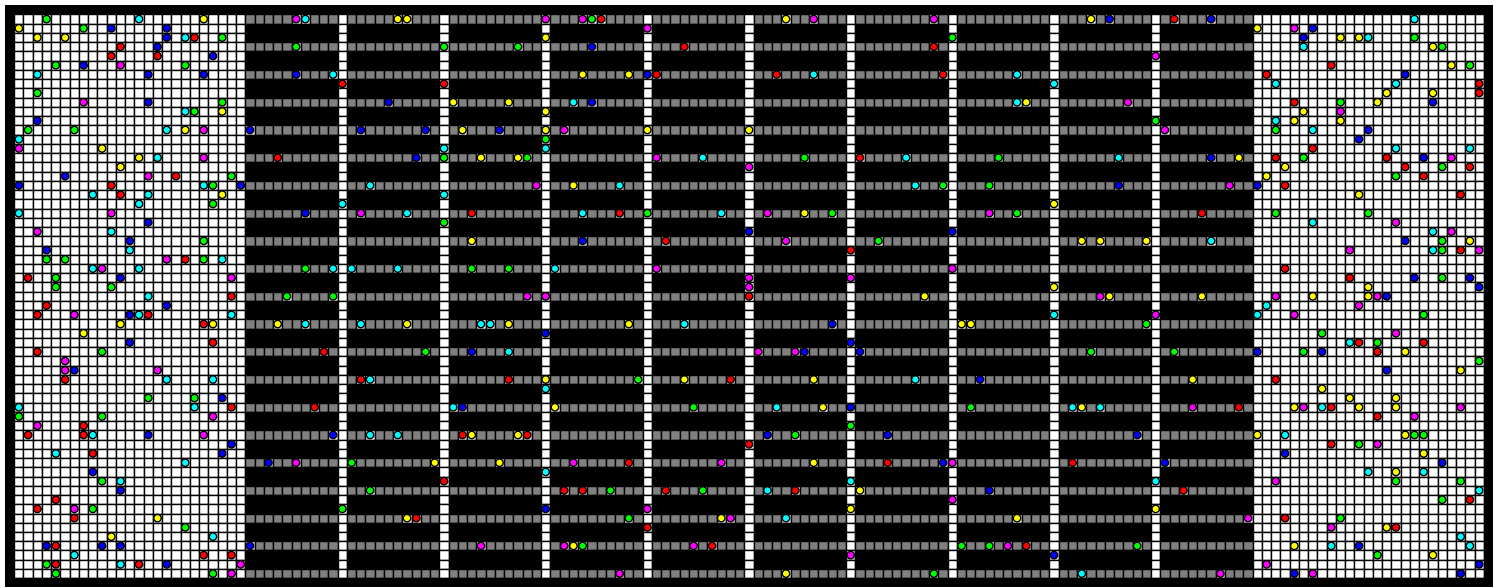}
        \caption{Warehouse-10-20-10-2-1 Map}
        \label{fig:warehouse1}
        \vspace{0.3cm}
    \end{subfigure}
    \begin{subfigure}{0.49\linewidth}
        \centering
        \includegraphics[width=0.6\linewidth]{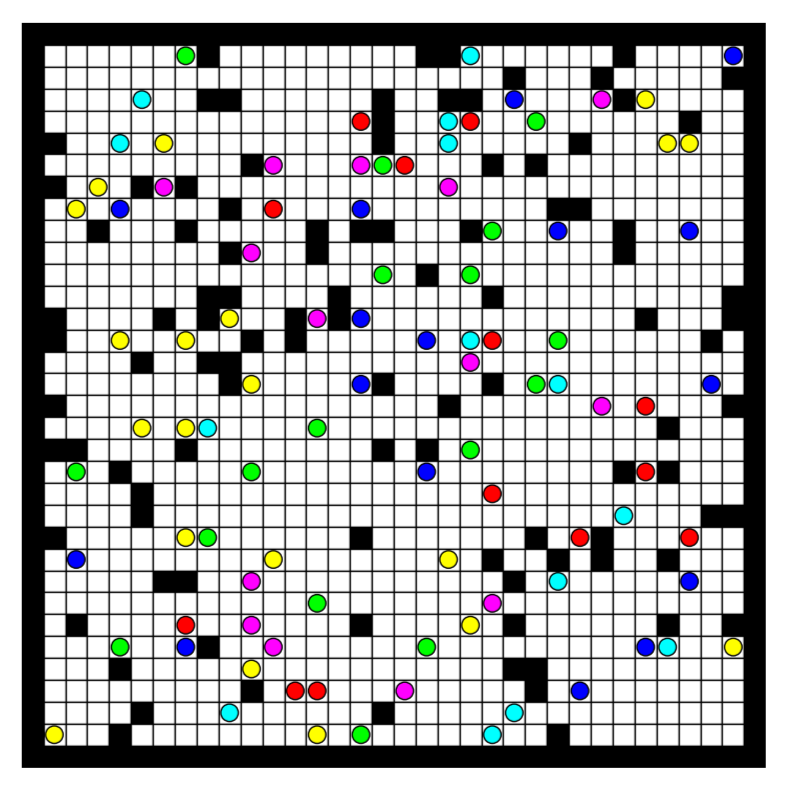}
        \caption{Random-32-32-10 Map}
        \label{fig:random}
    \end{subfigure}
    \begin{subfigure}{0.49\linewidth}
        \centering
        \includegraphics[width=0.8\linewidth]{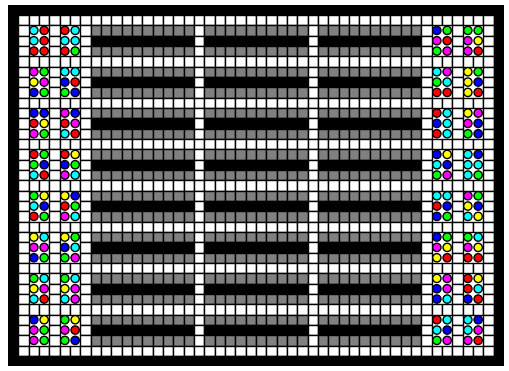}
        \caption{Kiva Warehosue Map}
        \label{fig:warehouse2}
    \end{subfigure}
    \caption{These figures represent warehouse environments for the MAPF experiments. (a) is a $63 \times 161$ 4-connected grid with 500 agents, (b) is a $32 \times 32$ 4-connected grid with 100 agents, (c) is a $33 \times 46$ 4-connected grid with 190 agents. Black cells are blocked. Gray cells are task endpoints. Colored circles are the initial locations of agents.}
    \vskip -0.1in
\end{figure*}

\textbf{Property 3.}
All tasks are eventually assigned and executed. \\ \\
\textit{Proof.}
At each timestep, Neural ATTF considers all unexecuted tasks for assignment, regardless of their current assignment status. Delayed agents are prioritized using a heuristic-based assignment module to preserve their assigned tasks. Since all unassigned tasks and idle agents are reconsidered at every timestep, and the number of tasks and agents is finite, each task is eventually assigned to an agent. Once assigned, successful execution follows from Property 1. Agents that complete tasks or reach the end of their planned paths are re-evaluated for new assignments, enabling continual progress.\\ \\
\textbf{Theorem 1.}
All well-formed MAPD instances are solvable, and Neural ATTF solves them.\\ \\
\textit{Proof.}
From Properties 1 and 2, agents can always find valid paths for executing tasks or safely idling. From Property 3, all tasks are eventually assigned and executed. Because agents plan sequentially with full path awareness and the algorithm ensures that no agent indefinitely blocks critical locations, the system avoids indefinite deadlocks and guarantees progress. Thus, Neural ATTF is complete for all well-formed MAPD instances.

\section{Experimental Results}

We conduct a comprehensive set of experiments to evaluate both the effectiveness of the Neural Space-Time A* path planner and the overall Neural ATTF algorithm. All experiments are performed over 25 independent instances, and the reported evaluation metrics are averaged across these runs to ensure statistical robustness. We use the following abbreviations throughout the results: ``st'' denotes the average service time per task, ``rt'' denotes the average runtime (ms) per timestep, ``tp'' denotes the task throughput, and ``iters'' denotes number of explored states in A* search in millions.

\begin{table}[t]
\caption{Comparison of windowed PBS with path planner of Neural ATTF }
\label{tab:pathplanner}\vskip 0.1in
\centering
\begin{tabular}{ccrrrr}
\toprule
\multirow{2}{*}{Map} & \multirow{2}{*}{Agents} & \multicolumn{2}{c}{wPBS} & \multicolumn{2}{c}{Nerual ATTF}  \\
\cline{3-6}
& & \multicolumn{1}{c}{tp} & \multicolumn{1}{c}{rt} & \multicolumn{1}{c}{tp} & \multicolumn{1}{c}{rt} \\
\midrule
\multirow{4}{*}{Random-32-32-10} &50 & 2.30 & \textbf{0.96} & \textbf{2.30} & 23.92\\
&100 & 4.36 & \textbf{5.77} & \textbf{4.46} & 40.46\\
&200 & \textbf{8.76} & \textbf{75.15} & 8.65 & 91.89\\
&300 & - & - & \textbf{8.88} & \textbf{322.96}\\
\midrule
\multirow{4}{*}{Warehouse-10-20-10-2-1} &200 & \textbf{3.19} & \textbf{27.35} & 3.09 & 57.50 \\
&300 & 4.13 & 146.95 & \textbf{4.22} & \textbf{137.36} \\
&400 & - & - & \textbf{5.04} & \textbf{300.41} \\
&500 & - & - & \textbf{5.38} & \textbf{691.39} \\
\midrule
\multirow{4}{*}{Kiva Warehouse}&130 & \textbf{4.33} & 135.02 & 4.30 & \textbf{63.83} \\
&150 & 4.75 & 241.82 & \textbf{4.85} & \textbf{78.00} \\
&170 & 5.20 & 398.22 & \textbf{5.34} & \textbf{96.15} \\
&190 & 5.58 & 628.22 & \textbf{5.79} & \textbf{155.71} \\
\bottomrule
\end{tabular}\vskip 0.1in
\end{table}

\subsection{Scalability Analysis of Neural STA*}

First, we evaluate the effectiveness of the path planner by replacing the PGTM Module with a random task assigner. The experiment is conducted on three different maps: a large warehouse environment of dimensions $63 \times 161$ containing shelves of size $10 \times 2$, arranged in a grid of $10 \times 20$ (Fig. \ref{fig:warehouse1}), a randomly generated map of dimensions $32 \times 32$ with $10\%$ obstacle coverage (Fig. \ref{fig:random}), and a smaller warehouse of dimensions $33 \times 46$ with shelves of size $10 \times 1$, arranged in a grid of $8 \times 3$ (Fig. \ref{fig:warehouse2}). Drop-off points are positioned on both sides of the shelves, and all pathways between the shelves are narrow, preventing two robots from passing in opposite directions. We compare our proposed algorithm to the state-of-the-art Windowed Priority-Based Search (Windowed PBS), which employs the Rolling-Horizon Collision Resolution (RHCR) technique for efficient path planning. In both methods, task assignment is performed randomly, meaning that whenever a robot completes a task, it is assigned a new start and goal location at random. We perform the experiment for 5000 timesteps, and then measure the following metrics:
\begin{itemize}
    \item Throughput ($tp$): Average the number of tasks completed per timestep across the experiment.
    \item Runtime per timestep ($rt$): Average time required per timestep to compute paths for all agents.
\end{itemize} 
The experiment times out if it takes more than 80 minutes, indicating that the algorithm cannot be used for real-time operations. The result of the experiment is shown in table \ref{tab:pathplanner}.

As seen from the table our algorithm demonstrates the capability to plan for a higher number of agents compared to wPBS across all tested maps, while achieving comparable or better performance. In the random map, the throughput of both algorithms remains nearly identical. However, wPBS fails to compute paths for 300 or more agents within the given time limit. A similar trend is observed in Warehouse 1, where wPBS is unable to generate plans for 400 or more agents. In contrast, Neural ATTF successfully plans for up to 500 agents while maintaining real-time performance. In Warehouse 2, Neural ATTF consistently outperforms wPBS in terms of both throughput and runtime, further demonstrating its scalability and efficiency.

\begin{table}[t]
\caption{Comparison of Neural STA* with baseline planner STA* }
\label{tab:experiment3}\vskip 0.1in
\centering
\resizebox{\linewidth}{!}{
\begin{tabular}{crrrrrr}
\toprule
\multirow{2}{*}{Agents} & \multicolumn{3}{c}{ST A*} & \multicolumn{3}{c}{Nerual ST A*}  \\
\cline{2-7}
& \multicolumn{1}{c}{st} & \multicolumn{1}{c}{rt} & \multicolumn{1}{c}{iters} & \multicolumn{1}{c}{st} & \multicolumn{1}{c}{rt} & \multicolumn{1}{c}{iters} \\
\midrule
100 & 332.27 & \textbf{30.00} & 0.32 & \textbf{329.27} & 98.61 & \textbf{0.22}\\
200 & \textbf{184.24} & \textbf{92.84} & 0.65 & 184.30 & 144.46 & \textbf{0.33}\\
300 & 138.14 & \textbf{190.41} & 1.05 & \textbf{137.32} & 208.63 & \textbf{0.44}\\
400 & \textbf{115.59} & 331.98 & 1.60 & 115.72 & \textbf{275.02} & \textbf{0.55}\\
500 & \textbf{102.21} & 532.46 & 2.27 & 103.56 & \textbf{376.54} & \textbf{0.67}\\
\bottomrule
\end{tabular}}
\end{table}

\begin{figure}[h]
    \centering
    \begin{subfigure}{\linewidth}
        \centering
        \includegraphics[width=0.85\linewidth]{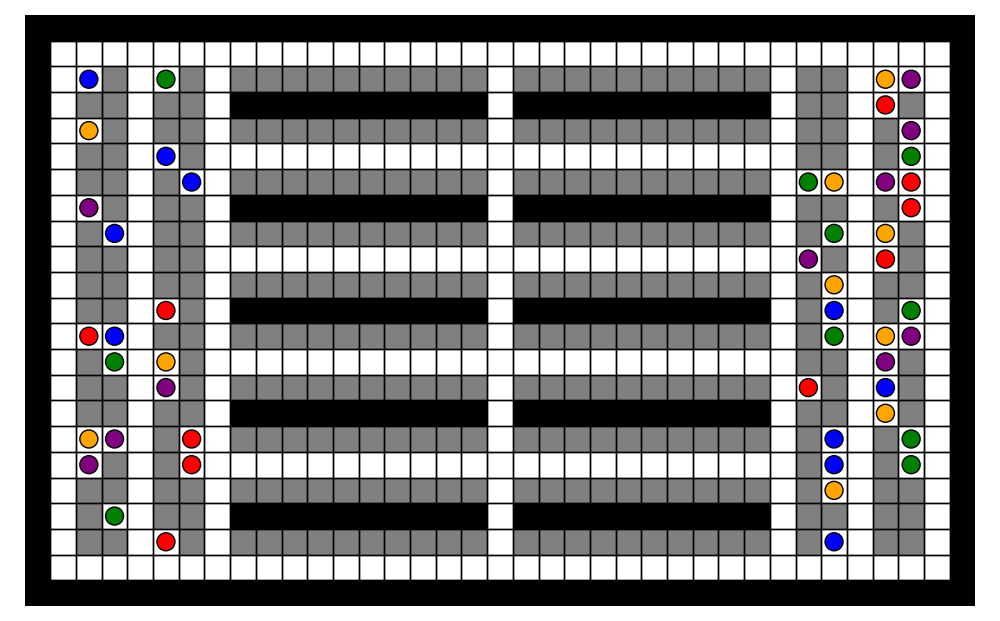}
        \caption{Small Warehouse}
        \label{fig:small-exp1}
    \end{subfigure}
    \begin{subfigure}{\linewidth}
        \centering
        \includegraphics[width=0.85\linewidth]{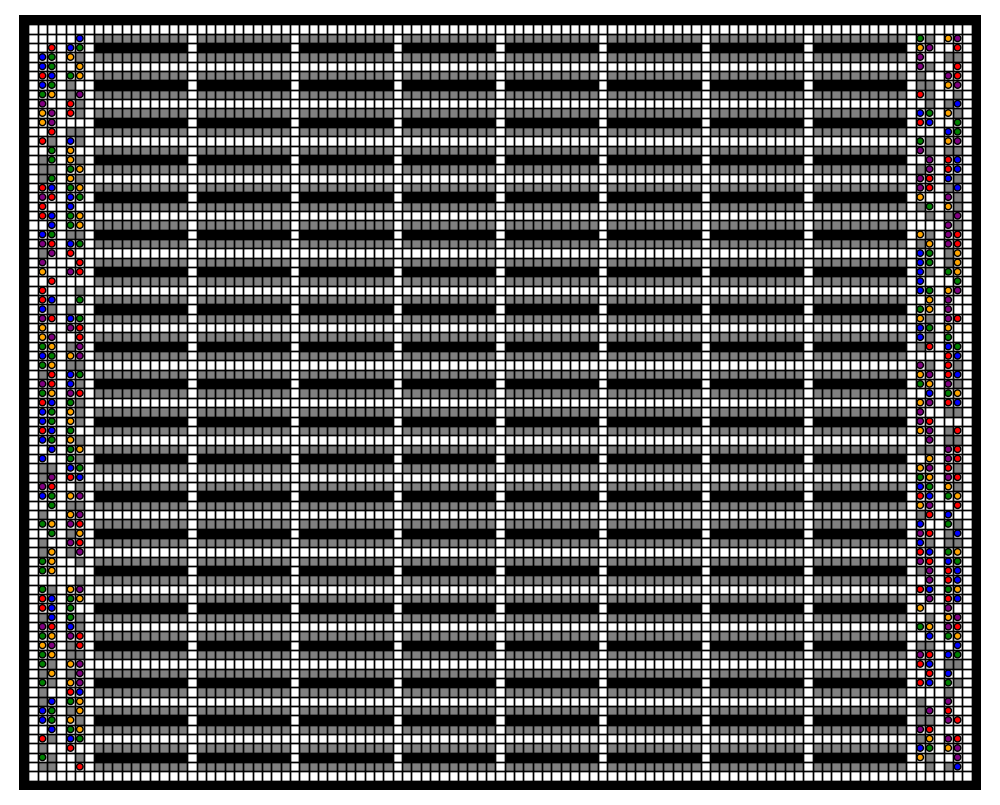}
        \caption{Large Warehouse}
        \label{fig:scalable}
    \end{subfigure}
    \caption{These figures represent warehouse environments for the MAPD experiments. (a) is a $35 \times 21$ 4-connected grid with 50 agents, (b) is an $101 \times 81$ 4-connected grid with 500 agents. Black cells are blocked. Gray cells are task endpoints. Colored circles are the initial locations of agents.}
    \vskip -0.1in
\end{figure}

\subsection{Efficiency Analysis of Neural STA*}

Next, to evaluate the effectiveness of the Neural Space-Time A* module, we compare its performance with a Neural Space-Time A* (Neural STA*) planner and a simple Space-Time A* (STA*) planner on the large warehouse map (figure \ref{fig:scalable}) for high task frequency (50) with varying numbers of agents (100, 200, 300, 400, 500). We assess the following metrics:
\begin{itemize}
    \item Average service time ($st$) - Average time taken to complete a task since its arrival
    \item Runtime per timestep ($rt$)
    \item Total number of expanded nodes across all searches, in millions ($iters$)
\end{itemize} 
The results of this experiment are presented in Table \ref{tab:experiment3}. 

The results indicate that service times remain consistent, with differences staying within $1\%$, confirming that the transition to Neural STA* does not compromise performance. However, Neural STA* significantly reduces the number of A* search iterations, with reductions ranging from $31\%$ to $70\%$ as the number of agents increases. For low agent counts (100, 200, 300), the encoder's inference time dominates, masking the benefits of reduced search iterations. In contrast, for higher agent counts (400, 500), the impact of Neural STA* becomes evident, leading to a reduction in runtime by $17.2\%$ and $29.3\%$ for 400 and 500 agents, respectively. These results highlight Neural STA*'s ability to improve computational efficiency without sacrificing task performance, particularly in scenarios with a large number of agents. Consequently, we adopt Neural STA* as the low-level planner for further experiments.

\begin{table*}[t]
\small
\centering
\caption{Results on MAPD instances in small warehouse}
\label{tab:experiment1}
\vskip 0.1in
% \scriptsize
\resizebox{0.92\linewidth}{!}{
\renewcommand{\arraystretch}{1.15}
\begin{tabularx}{\textwidth}{cc>{\raggedleft\arraybackslash}X>{\raggedleft\arraybackslash}X>{\raggedleft\arraybackslash}X>{\raggedleft\arraybackslash}X>{\raggedleft\arraybackslash}X>{\raggedleft\arraybackslash}X>{\raggedleft\arraybackslash}X>{\raggedleft\arraybackslash}X>{\raggedleft\arraybackslash}X>{\raggedleft\arraybackslash}X>{\raggedleft\arraybackslash}X>{\raggedleft\arraybackslash}X}
\toprule
\multirow{2}{*}{f} & \multirow{2}{*}{Agents} & \multicolumn{2}{c}{TPTS} & \multicolumn{2}{c}{CENTRAL} & \multicolumn{2}{c}{RMCA} & \multicolumn{2}{c}{LNS-PBS} & \multicolumn{2}{c}{LNS-wPBS} & \multicolumn{2}{c}{Neural ATTF} \\
\cline{3-14}
 & & \multicolumn{1}{c}{st} & \multicolumn{1}{c}{rt} & \multicolumn{1}{c}{st} & \multicolumn{1}{c}{rt} & \multicolumn{1}{c}{st} & \multicolumn{1}{c}{rt} & \multicolumn{1}{c}{st} & \multicolumn{1}{c}{rt} & \multicolumn{1}{c}{st} & \multicolumn{1}{c}{rt} & \multicolumn{1}{c}{st} & \multicolumn{1}{c}{rt}\\
\midrule
0.2 & 10 & 29.18 & \textbf{5.18}  & 29.77 & 28.16  & 28.00 & 198.33 & 27.92 & 313.45 & \textbf{27.87} & 316.08 & 28.98 & 8.15 \\
    & 20 & 25.75 & 29.35  & 26.70 & 136.21 & \textbf{24.98} & 198.72 & 25.33 & 294.94 & 25.67 & 316.50 & 25.73 & \textbf{9.07} \\
    & 30 & 24.46 & 68.80 & 25.56 & 305.78 & \textbf{24.08} & 198.94 & 25.10 & 292.04 & 24.69 & 298.29 & 24.78 & \textbf{9.58} \\
    & 40 & 23.75 & 134.16 & 25.46 & 415.25 & \textbf{23.23} & 199.55 & 24.30 & 286.58 & 24.58 & 291.88 & 24.21 & \textbf{10.21} \\
    & 50 & 23.37 & 196.81 & 25.05 & 757.40 & \textbf{22.98} & 200.05 & 24.09 & 277.72 & 24.39 & 292.96 & 23.75 & \textbf{11.00} \\
\midrule
0.5 & 10 & 128.66 & \textbf{0.95} & 109.71 & 51.23 & 104.08 & 438.57 & 116.59 & 400.44 & 117.44 & 382.50 & \textbf{98.50} & 9.70 \\
    & 20 & 30.92  & 26.68 & 27.99 & 172.36  & \textbf{26.25}  & 496.28 & 26.91  & 646.19 & 27.52  & 617.43 & 28.85  & \textbf{12.20} \\
    & 30 & 27.77  & 82.42 & 26.23 & 512.04  & \textbf{24.27}  & 501.24 & 25.26  & 667.61 & 25.72  & 635.44 & 26.22  & \textbf{12.76} \\
    & 40 & 26.32  & 141.32 & 25.39 & 1017.49 & \textbf{23.32}  & 503.49 & 24.65  & 667.25 & 24.84  & 657.18 & 25.19  & \textbf{13.73} \\
    & 50 & 25.79  & 235.59 & 24.94 & 1736.70 & \textbf{23.04}  & 508.45 & 23.94  & 666.01 & 24.76  & 645.86 & 24.51  & \textbf{14.93} \\
\midrule
1 & 10 & 311.66 & \textbf{0.92} & 285.75 & 65.70 & 267.30 & 464.67 & 273.48 & 448.81 & 266.77 & 419.59 & \textbf{213.07} & 10.12 \\
  & 20 & 95.44 & \textbf{6.21} & 75.13 & 266.76 & \textbf{62.44} & 851.97 & 67.21 & 880.60 & 67.20 & 762.55 & 64.83 & 12.94 \\
  & 30 & 43.21 & 26.70 & 31.41 & 492.12 & \textbf{26.85} & 974.76 & 28.82 & 1030.93 & 28.05 & 947.47 & 27.00 & \textbf{13.64} \\
  & 40 & 33.27 & 208.33 & 28.33 & 1381.56 & \textbf{24.36} & 987.04 & 25.28 & 1042.05 & 25.62 & 960.76 & 25.49 & \textbf{17.77} \\
  & 50 & 31.32 & 446.73 & 27.38 & 3238.17 & \textbf{23.56} & 995.00 & 24.42 & 1055.77 & 25.30 & 958.01 & 24.08 & \textbf{18.94} \\
\midrule
2 & 10 & 418.98 & \textbf{0.90} & 388.21 & 81.35 & 371.27 & 231.32 & 361.59 & 258.75 & 356.90 & 229.33 & \textbf{299.36} & 10.86 \\
  & 20 & 183.93 & \textbf{5.20} & 162.00 & 424.18 & 146.81 & 444.33 & 140.27 & 477.73 & 140.22 & 420.59 & \textbf{119.52} & 13.04 \\
  & 30 & 112.96 & 21.15 & 85.89 & 702.22 & 77.75 & 635.75 & 75.45 & 749.36 & 74.30 & 597.24 & \textbf{70.24} & \textbf{15.12} \\
  & 40 & 79.44 & 57.90 & 57.53 & 1440.20 & 43.49 & 798.31 & 44.55 & 1307.76 & 41.90 & 752.98 & \textbf{38.90} & \textbf{17.11} \\
  & 50 & 62.17 & 133.93 & 41.43 & 2206.70 & 28.88 & 927.25 & 30.46 & 1249.02 & 28.30 & 893.02 & \textbf{25.23} & \textbf{18.92} \\
\midrule
5 & 10 & 487.98 & \textbf{0.89} & 455.16 & 85.32 & 435.70 & 99.57 & 412.75 & 157.27 & 408.77 & 109.84 & \textbf{396.88} & 11.75 \\
  & 20 & 254.94 & \textbf{5.79} & 229.55 & 422.41 & 209.55 & 184.11 & 197.28 & 244.39 & 197.51 & 187.50 & \textbf{180.68} & 14.59 \\
  & 30 & 171.18 & 24.35 & 147.76 & 1012.82 & 132.06 & 268.07 & 126.41 & 373.18 & 123.95 & 272.18 & \textbf{115.45} & \textbf{16.83} \\
  & 40 & 131.09 & 64.82 & 108.28 & 1745.05 & 96.81 & 362.65 & 90.01 & 627.75 & 91.01 & 364.22 & \textbf{84.13} & \textbf{19.14} \\
  & 50 & 111.72 & 132.79 & 86.90 & 2686.08 & 74.32 & 425.26 & 70.31 & 914.49 & 72.25 & 422.82 & \textbf{67.39} & \textbf{22.17} \\
\midrule
10 & 10 & 509.36 & \textbf{0.85} & 478.17 & 92.96 & 458.23 & 56.68 & 438.71 & 117.76 & 431.76 & 65.62 & \textbf{427.24} & 12.30 \\
  & 20 & 269.21 & \textbf{5.30} & 242.18 & 375.23 & 228.90 & 101.20 & 217.33 & 168.63 & 215.74 & 110.00 & \textbf{211.84} & 16.69 \\
  & 30 & 190.99 & 19.64 & 165.13 & 869.85 & 154.28 & 152.34 & 146.56 & 254.68 & 144.11 & 163.94 & \textbf{141.03} & \textbf{19.42} \\
  & 40 & 150.80 & 51.94 & 128.39 & 1723.10 & 115.04 & 208.32 & 110.41 & 381.89 & 109.10 & 203.33 & \textbf{103.03} & \textbf{21.78} \\
  & 50 & 129.28 & 120.48 & 106.70 & 7442.20 & 94.29 & 246.96 & 88.75 & 602.85 & 89.33 & 243.87 & \textbf{84.13} & \textbf{22.18} \\
\bottomrule
\end{tabularx}}
\vskip -0.1in
\end{table*}

\subsection{Neural ATTF vs. State-of-the-Art}

We then compare Neural ATTF with various state-of-the-art algorithms: TPTS, CENTRAL, RMCA, LNS-PBS, LNS-wPBS, and HBH-MLA*. First, we use a small warehouse environment (figure \ref{fig:small-exp1}). It is a 4-connected grid of size $35 \times 21$, which consists of 4 columns of task endpoints on each side of the map, $2 \times 5$ block of shelves, and task endpoints on either sides of the shelves. The initial position of the agents are the non task endpoints. We generate a sequence of 500 tasks, with start and goal locations randomly selected from the task endpoints. The algorithms are tested under varying task frequencies (0.2, 0.5, 1, 2, 5, and 10) and different numbers of agents (10, 20, 30, 40, and 50). To evaluate the algorithms, we compare their service time ($st$) (as a measure of effectiveness) and runtime per timestep ($rt$) (as a measure of computational efficiency) for TPTS, CENTRAL, RMCA, LNS-PBS, LNS-wPBS, and Neural ATTF. We don't compare the results of HBH-MLA* since its service times are worse than TPTS and CENTRAL. The results of this experiment are presented in Table \ref{tab:experiment1}.

For lower task frequencies (0.2, 0.5, 1), RMCA achieved the best service time among all algorithms. However, Neural ATTF yielded an average of only 5\% higher service time than RMCA across all agent counts for these frequencies while demonstrating remarkable computational efficiency, reducing runtime by over 94\% on average compared to all other algorithms for all low frequency cases. The high runtimes of CENTRAL, RMCA, LNS-PBS, and LNS-wPBS, especially in high agent scenarios, hinder their suitability for real-time operations. Neural ATTF matched the service times of TPTS and CENTRAL for a task frequency of 0.2 and outperformed them with an average improvement of 8\% for a task frequency of 0.5 and 29\% for a task frequency of 1. For high task frequencies (2, 5, 10), Neural ATTF dominated in both service time and runtime. An exception occurred with low agent counts, where TPTS had a slightly lower runtime than Neural ATTF, though the difference was negligible for practical applications. Neural ATTF achieved superior service times, with an average improvement of 33\% over TPTS, 21\% over CENTRAL, 11.5\% over RMCA, 8\% over LNS-PBS, and 6\% over LNS-wPBS. Although the improvement over LNS-PBS and LNS-wPBS in service time is moderate, the significant reduction in computational runtime makes Neural ATTF a clear choice over them. The high runtimes of CENTRAL, LNS-PBS and LNS-wPBS, especially for large agent counts, make its real-time usage impractical. On average, Neural ATTF maintained runtimes over 90\% lower than other algorithms (excluding TPTS) across all agent counts and high frequencies. 

\begin{table*}[t]
\caption{Results on MAPD instances in large warehouse. ``N/A" means the algorithm failed to produce a solution in 1.5h.}
\label{tab:experiment2}\vskip 0.1in
\centering
\resizebox{0.8\linewidth}{!}{
\begin{tabular}{crrrrrrrrrr}
\toprule
\multirow{2}{*}{Agents} & \multicolumn{2}{c}{HBH+MLA*} & \multicolumn{2}{c}{RMCA} & \multicolumn{2}{c}{LNS-wPBS} & \multicolumn{2}{c}{LNS-PBS} & \multicolumn{2}{c}{Neural ATTF} \\
\cline{2-11}
& \multicolumn{1}{c}{st} & \multicolumn{1}{c}{rt} & \multicolumn{1}{c}{st} & \multicolumn{1}{c}{rt} & \multicolumn{1}{c}{st} & \multicolumn{1}{c}{rt} & \multicolumn{1}{c}{st} & \multicolumn{1}{c}{rt} & \multicolumn{1}{c}{st} & \multicolumn{1}{c}{rt}\\
\midrule
100 & 362.70 & \textbf{1.99} & 329.58 & 565.76 & \textbf{300.90} & 87.35 & 301.78 & 345.36 & 329.27 & 98.61\\
200 & 207.76 & \textbf{6.75} & 192.67 & 2,072.98 & 176.81 & 220.28 & \textbf{176.13} & 3,065.95 & 184.30 & 144.46\\
300 & 157.11 & \textbf{14.89} & 147.42 & 4,734.94 & 139.33 & 465.78 & 137.97 & 8,844.98 & \textbf{137.32} & 208.63\\
400 & 136.40 & \textbf{32.59} & 126.44 & 9,906.40 & 123.32 & 806.54 & N/A & N/A & \textbf{115.72} & 275.02\\
500 & 125.42 & \textbf{65.79} & N/A & N/A & 113.78 & 1,385.90 & N/A & N/A & \textbf{103.56} & 376.54\\
\bottomrule
\end{tabular}}
\end{table*}

\subsection{Scalability Analysis of Neural ATTF}

In the next experiment, we test the scalability of our algorithm. For this we use a larger warehouse environment (figure \ref{fig:scalable}). It is a 4-connected grid of size $101 \times 81$, which consists of 4 columns of task endpoints on each side of the map, $8 \times 40$ block of shelves, and task endpoints on either sides of the shelves. Again, the initial positions of the agents are the only non task endpoints. In this experiment, we generate 1,000 tasks in a manner similar to the previous experiment and evaluate the algorithms with a task frequency of 50 tasks per timestep. The tests are conducted with varying numbers of agents: 100, 200, 300, 400, and 500. We evaluate algorithms by comparing service time ($st$) and runtime per timestep ($rt$) for HBH-MLA*, RMCA, LNS-PBS, LNS-wPBS, and Neural ATTF. The results to this experiment are given in table \ref{tab:experiment2}.

Among the tested algorithms, HBH-MLA* exhibited the lowest runtime but at a significant performance cost, with service times averaging $18.4\%$ higher than the best-performing algorithms in similar scenarios. Although runtime is not the primary focus of this study due to hardware dependencies, Neural ATTF maintains runtimes well within the threshold required for real-time operations, assuming a threshold of 1 second per timestep \cite{ma2017lifelong}. For low agent counts (100, 200), LNS-wPBS demonstrated strong performance, achieving competitive service times with a balanced runtime. However, RMCA and LNS-PBS incurred extremely high runtimes that grew non linearly with increasing agent counts, making them unsuitable for real-time applications. Neural ATTF outperformed LNS-wPBS in runtime, with service times only $6\%$ higher on average for low agent counts. For agent counts exceeding 200, Neural ATTF consistently achieved the best service times among all algorithms. In contrast, RMCA and LNS-PBS either failed to produce solutions within the 1.5-hour time limit or had prohibitively high runtimes. Additionally, LNS-wPBS exhibited more than a $100\%$ increase in runtime compared to Neural ATTF for higher agent counts. 

\subsection{Robustness Evaluation Under Agent Execution Delays}

Finally, to show the robustness of the algorithm, we test the algorithm for different delay probabilities of 0 \%, 1 \%, and 2 \%. Here delay probability is the probability with which an agent fails to perform a planned action. This experiment was conducted on the large warehouse map (figure \ref{fig:scalable}) with varying number of agents (100, 200, 300, 400, 500) and a task frequency of 50 tasks per timestep. A 1\% delay probability corresponds to approximately 5 agents failing to execute their planned actions per timestep for 500 agents, while a 2\% delay probability results in about 10 agents failing per timestep in the same scenario. We compare the service time ($st$) and runtime per timestep ($rt$) to judge the performance of the algorithm in each case. The results to this experiment are given in table \ref{tab:experiment4}.

\begin{table}[t]
\caption{Comparison under varying delay probabilities, $p$}
\label{tab:experiment4}\vskip 0.1in
\centering
\resizebox{\linewidth}{!}{
\begin{tabular}{crrrrrr}
\toprule
\multirow{2}{*}{Agents} & \multicolumn{2}{c}{$p=0$} & \multicolumn{2}{c}{$p=0.01$} & \multicolumn{2}{c}{$p=0.02$} \\
\cline{2-7}
& \multicolumn{1}{c}{st} & \multicolumn{1}{c}{rt} & \multicolumn{1}{c}{st} & \multicolumn{1}{c}{rt} & \multicolumn{1}{c}{st} & \multicolumn{1}{c}{rt} \\
\midrule
100 & 329.27 & 98.61 & 337.32 & 106.52 & 340.90 & 116.08\\
200 & 184.30 & 144.46 & 188.88 & 217.47 & 191.66 & 211.26\\
300 & 137.32 & 208.63 & 140.28 & 363.22 & 144.16 & 397.77\\
400 & 115.72 & 275.02 & 118.72 & 580.17 &121.49 & 669.46\\
500 & 103.56 & 376.54 & 107.56 & 898.38 &110.56& 1057.31\\
\bottomrule
\end{tabular}}\vskip -0.1in
\end{table}

The results show that both service time and runtime increase as the delay probabilities rise. However, the service time remains robust, with a delay probability of $1\%$ resulting in a service time increase of only $3.5\%$ compared to no delay, and a delay probability of $2\%$ causing a $7\%$ increase. This indicates only a minor degradation in performance under uncertainty. The runtime, on the other hand, experiences a significant increase due to the added replanning overhead introduced by delays. For a delay probability of $1\%$, the algorithm roughly adds replanning time equivalent to handling five agents per timestep, while a $2\%$ delay probability increases this to approximately ten agents per timestep in scenario involving 500 agents. These findings demonstrate that while the algorithm effectively maintains its performance in uncertain environments, managing delays introduces additional computational costs primarily due to the need for frequent replanning.

\section{Conclusion and Future Work}

Neural ATTF consistently balances performance and computational efficiency, achieving excellent service times while maintaining significantly lower runtimes, making it ideal for real-time applications. Unlike RMCA, LNS-PBS, and LNS-wPBS, which face exponential runtime growth, Neural ATTF scales efficiently across diverse scenarios. While HBH-MLA* offers lower runtimes, its substantial trade-off in service time limits its practicality in high-demand settings. The integration of Neural STA* further enhances Neural ATTF’s computational efficiency, reducing A* search iterations by up to $70\%$ for larger agent counts without sacrificing performance. While the benefits are less evident for small agent counts due to encoder inference overhead, the improvements become significant for large-scale problems, ensuring scalability and adaptability. Neural ATTF also demonstrates robustness in uncertain conditions, with minimal service time degradation under higher delay probabilities. Although delays introduce additional replanning costs, Neural ATTF efficiently manages this overhead, maintaining reliable task completion despite disruptions.

In summary, Neural ATTF distinguishes itself as a versatile, high-performance algorithm that excels across a variety of operational settings. Its ability to balance service time and runtime, coupled with enhanced scalability through Neural STA* and resilience to environmental uncertainty, makes it an ideal choice for real-time multi-agent path planning in complex and dynamic environments. These characteristics position Neural ATTF as a leading solution for addressing the challenges of modern multi-agent systems, particularly in applications requiring high throughput, reliability, and computational efficiency.

Future work involves deploying Neural ATTF in a robotics simulator to evaluate its performance in realistic, dynamic scenarios and subsequently applying it to real-world tasks like warehouse automation and multi-agent coordination. 

% \nolinenumbers

\section*{Acknowledgments}
This study was supported by Hyundai-MOBIS under the grant number AWD-006126. This support does not constitute an endorsement by the funding agency of the opinions expressed in the paper.

% \bibliographystyle{IEEEtran} % or another style like plain, abbrv, etc.
% \bibliography{content} 

% Generated by IEEEtran.bst, version: 1.14 (2015/08/26)

\end{document}